\documentclass{article}

\usepackage{microtype}
\usepackage{graphicx}
\usepackage{subfigure}
\usepackage{subcaption}
\usepackage{booktabs} 
\usepackage[table]{xcolor}
\usepackage{amsmath}
\usepackage{amsfonts}
\usepackage{hyperref}
\usepackage{xcolor}
\usepackage{float}
\usepackage{multirow}
\usepackage{threeparttable}

\newcommand{\Ours}{\textsc{DAS}\xspace}

\usepackage[accepted]{{mlsys2026}}

\mlsystitlerunning{Beat the long tail: Distribution-Aware Speculative Decoding for RL Training}

\usepackage{xspace}

\NewDocumentCommand{\var}{O{s} m O{}}{%
  \ensuremath{#1_{#2}^{#3}}
}
\usepackage{siunitx}

\newcommand{\commentout}[1]{}

\definecolor{light-gray}{gray}{0.80}

\usepackage{amsthm}

\usepackage{amsthm}

\newtheorem{mytheorem}{Theorem}
\newtheorem{assumption}[mytheorem]{Assumption}
\newtheorem{lemma}[mytheorem]{Lemma}
\newtheorem{remk}[mytheorem]{Remark}

\usepackage{tikz}
\usetikzlibrary{shapes.misc}

\usepackage[skip=2pt]{caption}
\usepackage{setspace}
\usepackage{enumitem}
\usepackage{titlesec}

\setlist{nosep} 
\newlength{\subsectionbelowskip}
\newlength{\subsectionaboveskip}

\newlength{\paragraphaboveskip}

\newcommand{\setvspace}[2]{%
  #1 = #2
  \advance #1 by -1\parskip}

\titlespacing*{\subsection}{0pt}{\subsectionaboveskip}{\subsectionbelowskip}
\titlespacing*{\subsubsection}{0pt}{\subsectionaboveskip}{\subsectionbelowskip}
\titlespacing*{\paragraph}{0pt}{\paragraphaboveskip}{*}

\makeatletter 
\def\thm@space@setup{%
  \thm@preskip=3pt
  \thm@postskip=\thm@preskip 
}
\makeatother

\setlist[itemize]{noitemsep, topsep=0pt}

\makeatletter
\g@addto@macro\normalsize{%
  \setlength\abovedisplayskip{1pt}
  \setlength\belowdisplayskip{1pt}
  \setlength\abovedisplayshortskip{1pt}
  \setlength\belowdisplayshortskip{1pt}
}
\makeatother

\makeatletter 
\renewenvironment{proof}[1][\proofname]{\par
  \vspace{1pt}
  \pushQED{\qed}%
  \normalfont
  \topsep0pt \partopsep0pt 
  \trivlist
  \item[\hskip\labelsep
        \itshape
    #1\@addpunct{.}]\ignorespaces
}{%
  \popQED\endtrivlist\@endpefalse
  \addvspace{3pt plus 3pt} 
}
\makeatother 

\begin{document}

\twocolumn[
\mlsystitle{Beat the long tail: Distribution-Aware Speculative Decoding for RL Training}



\mlsyssetsymbol{equal}{*}

\begin{mlsysauthorlist}
\mlsysauthor{Zelei Shao}{equal,uiuc,together}
\mlsysauthor{Vikranth Srivatsa}{equal,together,ucsd}
\mlsysauthor{Sanjana Srivastava}{together}
\mlsysauthor{Qingyang Wu}{together}
\mlsysauthor{Alpay Ariyak}{together}
\mlsysauthor{Xiaoxia Wu}{together}
\mlsysauthor{Ameen Patel}{prime}
\mlsysauthor{Jue Wang}{together}
\mlsysauthor{Percy Liang}{together}
\mlsysauthor{Tri Dao}{together}
\mlsysauthor{Ce Zhang}{together}
\mlsysauthor{Yiying Zhang}{ucsd}
\mlsysauthor{Ben Athiwaratkun}{together}
\mlsysauthor{Chenfeng Xu}{together}
\mlsysauthor{Junxiong Wang}{together}
\end{mlsysauthorlist}

\mlsysaffiliation{together}{Together AI}
\mlsysaffiliation{uiuc}{University of Illinois Urbana-Champaign}
\mlsysaffiliation{ucsd}{University of California San Diego}
\mlsysaffiliation{prime}{Prime Intellect}

\mlsyscorrespondingauthor{Zelei Shao}{zelei2@illinois.edu}
\mlsyscorrespondingauthor{Vikranth Srivatsa}{vsrivatsa@ucsd.edu}

\mlsyskeywords{RL training, Speculative Decoding, MLSys}

\vskip 0.3in

\begin{abstract}
Reinforcement learning (RL) post-training has become essential for aligning large language models (LLMs), yet its efficiency is increasingly constrained by the rollout phase, where long trajectories are generated token by token. We identify a major bottleneck—the long-tail distribution of rollout lengths, where a small fraction of long generations dominates wall-clock time—and a complementary opportunity—the availability of historical rollouts that reveal stable prompt-level patterns across training epochs. Motivated by these observations, we propose \textbf{DAS, a Distribution-Aware Speculative decoding framework} that accelerates RL rollouts without altering model outputs. DAS integrates two key ideas: an \textbf{adaptive, nonparametric drafter} built from recent rollouts using an incrementally maintained suffix tree, and a \textbf{length-aware speculation policy} that allocates more aggressive draft budgets to long trajectories that dominate makespan. This design exploits rollout history to sustain acceptance while balancing base and token-level costs during decoding. Experiments on math and code reasoning tasks show that DAS reduces rollout time up to 50\% while preserving identical training curves, demonstrating that distribution-aware speculative decoding can significantly accelerate RL post-training without compromising learning quality.
\end{abstract}
]



\printAffiliationsAndNotice{\mlsysEqualContribution} 

\section{Introduction}

\label{sec:intro}

Reinforcement learning (RL) has emerged as a cornerstone of large language model (LLM) post-training. By directly optimizing for desired behaviors, RL enables models to align with human preferences and verifiable objectives through reward signals derived from task-specific feedback~\citep{ouyang2022training,bai2022constitutionalaiharmlessnessai}. 
With the success of in-production RL-trained models like DeepSeek-R1~\citep{guo2025deepseek}, we are seeing increasing interest and needs for RL training for large models and for more complex tasks like deep reasoning. 

As model size and context length grow, the computation bottleneck of RL training is shifting. Unlike classical fine-tuning, where the training (model weight updates) phase is the most time-consuming, modern RL training for LLMs is often dominated by the ``rollout'' phase, where the training dataset is tested with model inferencing. Our study, as well as other recent studies~\cite{verl_one_step_off_2024}, show that in practice, the rollout phase accounts for more than 70\% of the total training time, often exceeding the cost of backpropagation and parameter updates. The dominating rollout-phase time arises from a few factors: (1) the inherently autoregressive nature of LLM decoding, (2) increased LLM generation length~\citep{guo2025deepseek}, and (3) increasing sample size to achieve higher accuracy~\citep{wang2025reinforcement}.

Despite the dominance of the rollout phase, most existing RL systems are not designed to maximize rollout efficiency and instead allocate computing resources primarily for optimization or backpropagation phases~\cite{rlhfuse_2025}. A natural step to take to optimize the rollout phase is to borrow techniques from LLM serving systems, such as prefill-decode disaggregation~\citep{zhong2024distserve}, quantization~\citep{MLSYS2024_42a452cb}, and speculative decoding~\citep{fast_inference_via_spec_decoding}. 

Unfortunately, what works well for serving systems does not translate directly to the rollout phase in RL training for several reasons. \textit{Insight-1:} All samples in the training dataset need to complete their model inference before the next training phase can start. Today's serving systems optimize for Time-to-First-Token (TTFT) and Time-per-Output-Token (TPOT), resulting in long sequences taking more time to complete their generation and causing long tail latency in RL rollout phase. \textit{Insight-2:} The RL training process reuses the same set of samples in each training iteration, while LLM serving systems assume each user request is different. As a result, serving systems today do not leverage the nature of reappeared requests that could otherwise be used to improve RL rollout speed. \textit{Insight-3:} Unlike model serving where the LLMs are fixed, in RL training, the model weights keep getting updated. In such a dynamic environment, approaches that work in traditional serving may not work anymore.

An emerging set of works, some concurrent with our work, target to improve rollout efficiency~\citep{zhang2025fastgrpo, liu2025specrl, zhong2025streamrlscalableheterogeneouselastic}. While making progress, they do not fully solve the rollout bottleneck and do not properly leverage the three key rollout properties listed above.
Another line of work attempts to accelerate rollouts from the ML perspective by relaxing training fidelity constraints, for instance, through truncation or quantization strategies~\cite{zhong2025streamrlscalableheterogeneouselastic}. These methods often compromise learning stability or degrade performance in reasoning-heavy tasks where long-term rollouts are essential.

This paper proposes a comprehensive systems-ML codesign approach rooted from the unique properties of rollout in LLM RL training. Specifically, we build \textbf{\Ours{}}, an RL system that adapts traditional speculative decoding (SD) for RL rollout in three key different ways. 

First, we propose an \textit{adaptive, nonparametric drafter} method for drafting speculation tokens. From \textit{Insight-3}, during RL training, the model being trained (i.e., the target model) keeps changing, and using a pre-trained draft model would not work. Our proposal is to take advantage of the reuse nature (\textit{Insight-2}) to perform speculative decoding based on suffix-trees~\cite{oliaro2025suffixdecoding}. To remain aligned with the current policy, we prune the suffix structure dynamically, retaining only recent and relevant historical rollouts for improved locality and reliability.

Second, based on \textit{Insight-1}, we propose a distribution-aware speculative decoding method that allocates draft budget preferentially to high-difficulty, long-horizon prompts, those that dominate the rollout makespan rather than speculating uniformly. Instead of maintaining a single global suffix tree, we construct and update per-problem trees. This design better captures domain-specific patterns and improves the accuracy of speculation in heterogeneous tasks.

We implement \Ours{} on top of slightly modified versions of VeRL~\citep{sheng2024hybridflow} and vLLM~\citep{kwon2023efficient}, which are state-of-the-art RL training, speculative decoding implementation, and LLM inference engines, respectively. We evaluate the \Ours{} system on math~\citep{deepscaler2025} and code datasets~\citep{deepcoder} using up to six NVIDIA H100 GPU servers, with model sizes ranging from 1.5B to 8B parameters. Our results show that \Ours{} outperforms VeRL by up to 50\% in generation time, with no degradation in training accuracy.


Overall, this paper makes the following contributions:

\begin{itemize}
    \item Three key insights regarding the unique properties of the RL rollout phase and their system-level implications.
    \item \Ours, a full RL system that achieves up to 50\% rollout time reduction than the SoTA RL system.
    \item The proposal of a training-free, self-evolved speculative decoding method designed specifically for RL rollout.
    \item A distribution-aware speculative decoding method that effectively reduces tail latency in RL rollout.

\end{itemize}

\section{Related Works}
\label{sec:related}


\paragraph{RL post-training.} Field-standard RL post-training frameworks include VeRL~\cite{sheng2024hybridflow} and OpenRLHF~\cite{hu2024openrlhf}. RL post-training generally follows three phases: \emph{generation} of trajectories (autoregressive generation); \emph{preparation} of trajectories, such as reward labeling; and \emph{training}. These frameworks integrate various parallelism strategies, but the inference-heavy generation and preparation phases present opportunities for further performance gains through SD. 

We note two key discrepancies between RL post-training and serving workloads: (i) RL post-training is more structured, with the number of rollouts per iteration being known and the dataset being available \emph{a priori}; and (ii) because the dataset is revisited across epochs, it exposes \emph{global} and \emph{historical} signals about current rollout behavior. Using this information, our work aims to accelerate rollouts from a holistic, training-wide perspective.


\paragraph{Speculative decoding in throughput-oriented scenarios.}
Speculative decoding~\cite{leviathan2023fast,miao2023specinfer} accelerates autoregressive generation by trading additional computation for reduced latency. In throughput-oriented settings such as online serving, the primary objective is to maximize the number of accepted tokens per second under a fixed compute budget. Prior works have formulated this as an optimization problem over speculative configurations: Liu~\cite{liu2024optimizing} defines \emph{goodput} as the effective throughput and employs simulation-based search, while Huang~\cite{huang2025specserve} presents an analytical model to derive optimal draft and verify lengths.

However, these studies primarily target online-serving workloads and overlook properties unique to RL post-training.
First, they do not exploit the \emph{global} and \emph{historical} information available across RL training epochs. Second, they treat all requests as homogeneous,
whereas in synchronous on-policy RL training, longer generations dominate total latency while shorter ones contribute little. Our work identifies and models this imbalance, introducing differentiated optimization strategies for requests of varying lengths.



\paragraph{Online learning of drafters.}
Online SD adapts the drafter during use. \emph{Online Speculative Decoding} periodically distills the drafter on-the-fly from target-model corrections using spare FLOPs, reducing latency under distribution shift without growing the drafter~\citep{liu2024onlinesd}. \emph{Self-speculative} approaches eliminate external drafters by reusing the target model with layer skipping; later work (\emph{SWIFT}) makes this selection adaptive at run time, avoiding offline tuning~\citep{zhang2024selfsd,xia2024swift}. Model-free variants replace neural drafters with retrieval or symbolic structures—e.g., \emph{Prompt Lookup Decoding+} (PLD+) selects spans from the input and early-layer signals, while suffix-structure methods, SuffixDecoding \citep{oliaro2025suffixdecoding} builds a suffix tree ~\citep{weiner1973linear} over the tokens from the current and previous requests and uses it to generate speculative tokens. The root denotes the start of any stored suffix; each child represents a token that can follow its parent; and the path from the root to any node encodes a distinct subsequence.

Unlike these serving-oriented approaches, RL post-training presents a distinct adaptation challenge.
Prior online SD methods adjust to \emph{input}-distribution drift or varying system load,
whereas RL post-training features stable inputs but an evolving \emph{policy},
leading to staleness in neural drafters.
Conversely, repeated rollouts of the same problems
provide a unique opportunity for nonparametric, history-driven draft structures
that can be refreshed each epoch.
Our framework exploits this property to maintain high acceptance rates
and to schedule speculation into the idle GPU slack exposed during batched rollouts.

\paragraph{Speculative decoding for RL.} Some concurrent works have explored various uses of speculative decoding in RL post-training. SPEC-RL~\cite{liu2025specrl} uses prior trajectories as drafts, but introduces a lenience parameter for acceptance that changes the output distribution. Unlike \Ours, it does not recover non-SD-level accuracy. Furthermore, it does not consider differences in the contributions of various requests' to latency. FastGRPO~\cite{zhang2025fastgrpo} updates a neural draft model alongside the target model, consuming a considerable memory budget and compromising the scalability of the method. Finally, RhymeRL~\cite{he2025historyrhyme} takes advantage of trajectory similarity over time and considers batch-length skew. However, it lacks problem difficulty- and window-awareness. Problem difficulty-awareness is important for efficiently decoding valid trajectories and not token-wise similar invalid ones. Window-awareness is crucial for adapting to policy distribution shift, as trajectories from early policies lack similarity to those from later policies (Fig.~\ref{fig:generation_sim}). 






\begin{figure}[t]
    \centering
    \includegraphics[width=1\linewidth]{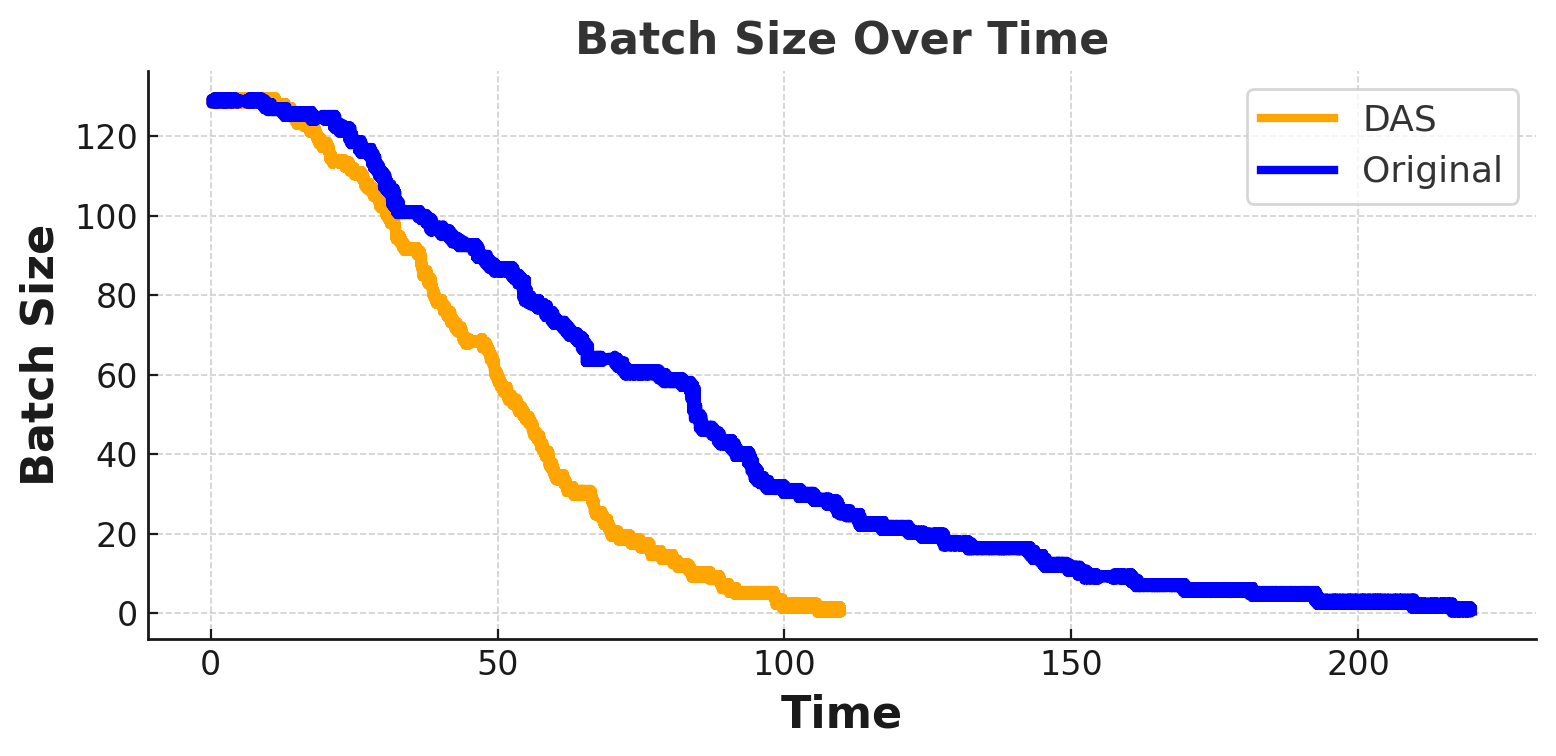}    \caption{\textbf{Effective batch size collapse during rollout w/o DAS.}
    Measured on the DeepSeek-distilled 7B model~\citep{guo2025deepseek} (DeepScaleR~\citep{deepscaler2025} prompts).
    As decoding progresses, short sequences finish first and the \emph{effective} batch size shrinks, leaving a few long stragglers to determine the step makespan. With our method, we can both reduce the total latency and alleviating the impact of long-tail stragglers.}
    \label{fig:batch_size_skew}
\end{figure}

\section{Rollouts in RL Training}

In this section, we present the unique characteristic of the RL workload.

\paragraph{Long-tail bottleneck in RL rollouts.}


RL post-training runs in iterations where the actor generates a fixed set of trajectories, and the learner updates the policy. In practice, systems like VeRL and OpenRLHF favor \emph{data-parallel} (DP) rollout workers to scale decoding throughput, resorting to \emph{tensor-parallelism} (TP) if the training model cannot fit into memory. 




Decoding begins at full parallelism, but as short sequences finish the \emph{effective} batch collapses and a few long responses (stragglers) determine the step makespan—classic long-tailed behavior also observed in LLM systems. This yields poor GPU utilization as workers idle while stragglers continue decoding; multiple RL/LLM studies report decoding dominating wall-clock time and under-utilizing accelerators despite sophisticated batching. Figure~\ref{fig:batch_size_skew} profiles the effective batch size running for a representative setup: after roughly 100 decode steps, the parallelism drops sharply, confirming the long-tail runtime bottleneck in RL training settings.


This motivates \emph{speculative decoding}~\citep{leviathan2023fast}: by drafting multiple tokens and verifying them in parallel, we can shrink straggler runtimes and reclaim idle GPU time, yielding substantial end-to-end speedups without changing model outputs. 
While speculative decoding is typically evaluated in small-batch, latency-sensitive serving, RL training exhibits \emph{effective} batch collapse during decoding—short sequences complete first and stragglers dominate the step. This exposes idle capacity that speculation can utilize.

\paragraph{High Similarity of RL Rollouts to Recent History}

\begin{figure}[H]
    \centering
    \includegraphics[width=1\linewidth]{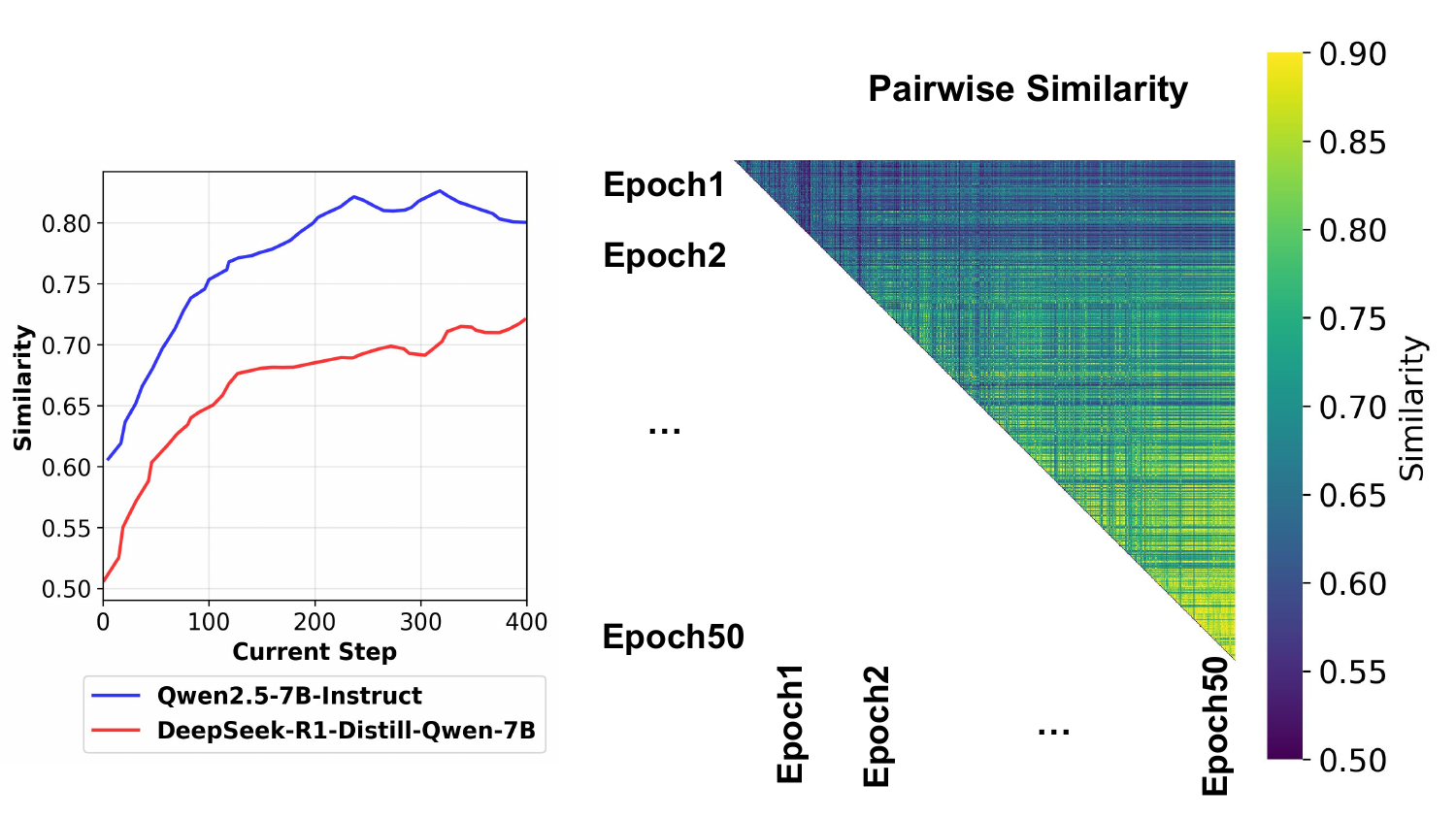}
    \caption{(\textbf{Left}) Content similarity per iteration using an N-gram to calculate this reuse ratio. (\textbf{Right}) Pairwise similarity across epochs for Qwen2.5-7B-Instruct. The block structure concentrated near the diagonal shows that rollouts are most similar to those from recent epochs, and similarity decays with temporal distance. 
    This reflects policy drift: as the policy is continually updated, older generations become less predictive of current behavior.}
    \label{fig:generation_sim}
\end{figure}

Across epochs, trajectories for the same prompts exhibit pronounced lexical and schematic reuse. In Figure~\ref{fig:generation_sim}, we observe elevated similarity with recent trajectories and reduced similarity with distant ones, consistent with policy drift. This recency bias implies that a history-indexed, model-free drafter (e.g., suffix array~\citep{manber1993suffix} or suffix tree~\citep{ukkonen1995line}) can mine recent continuations to propose high-quality drafts, even as the policy evolves.

\begin{figure*}
    \centering
    \includegraphics[width=.93\linewidth]{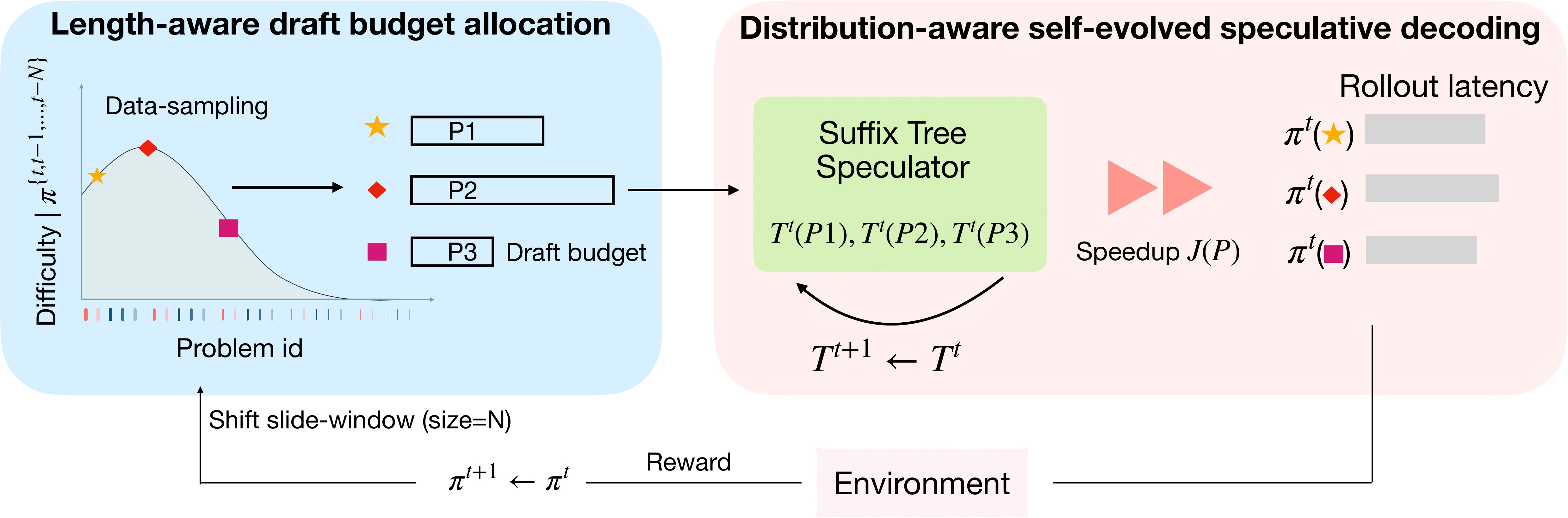}
    \caption{Overview of our rollout acceleration framework in RL training.
(\textbf{Left})~Length-aware draft budget allocation.
We estimate per-problem length from recent rollouts and assign a draft budget accordingly: problems predicted to be long or hard are allocated a more aggressive speculative budget, while easy problems receive little or no speculation.
This policy is updated over a sliding window of recent trajectories, so it adapts as the policy changes over training.
(\textbf{Right})~Distribution-aware, self-evolving speculative decoding.
For each problem shard, we maintain a suffix tree speculator that is incrementally updated from most recent rollouts. At decode time, the speculator proposes multi-token drafts drawn from high-frequency suffix matches, and the target model verifies them in parallel; accepted tokens advance generation with reduced rollout latency.}
    \label{fig:system_diag}
\end{figure*}

\section{Distribution-aware Speculative Decoding Framework}

We introduce a distribution-aware speculative decoding framework (\Ours), an RL post-training approach that accelerates rollout by adapting speculative decoding. Figure~\ref{fig:system_diag} provides an overview of the system. Building on RL pipelines that optimize a policy with preference- or verifiability-based rewards, \Ours{} maintains a history-indexed, nonparametric drafter that is continually refreshed from recent rollouts to preserve high acceptance as the policy evolves. The distribution-aware speculative decoding component then decides how much speculative budget to allocate to each request, favoring long, high-latency problems. By combining adaptive budget allocation with an online, distribution-aligned speculator (updated incrementally in the spirit of online suffix tree construction), the system reduces rollout latency without modifying the reward loop.

We motivate our choice of an adaptive, non-parametric drafting approach in Section~\ref{sec:training_free_drafter} and describe how to allocate the draft budget to reduce the runtime of long-tail sequences in Section~\ref{sec:dynamic_draft_budget}.
\subsection{Distribution-aware Drafter}
\label{sec:training_free_drafter}





\subsubsection{Parameterized static drafter}
\begin{figure}[ht]
    \centering    \includegraphics[width=1\linewidth]{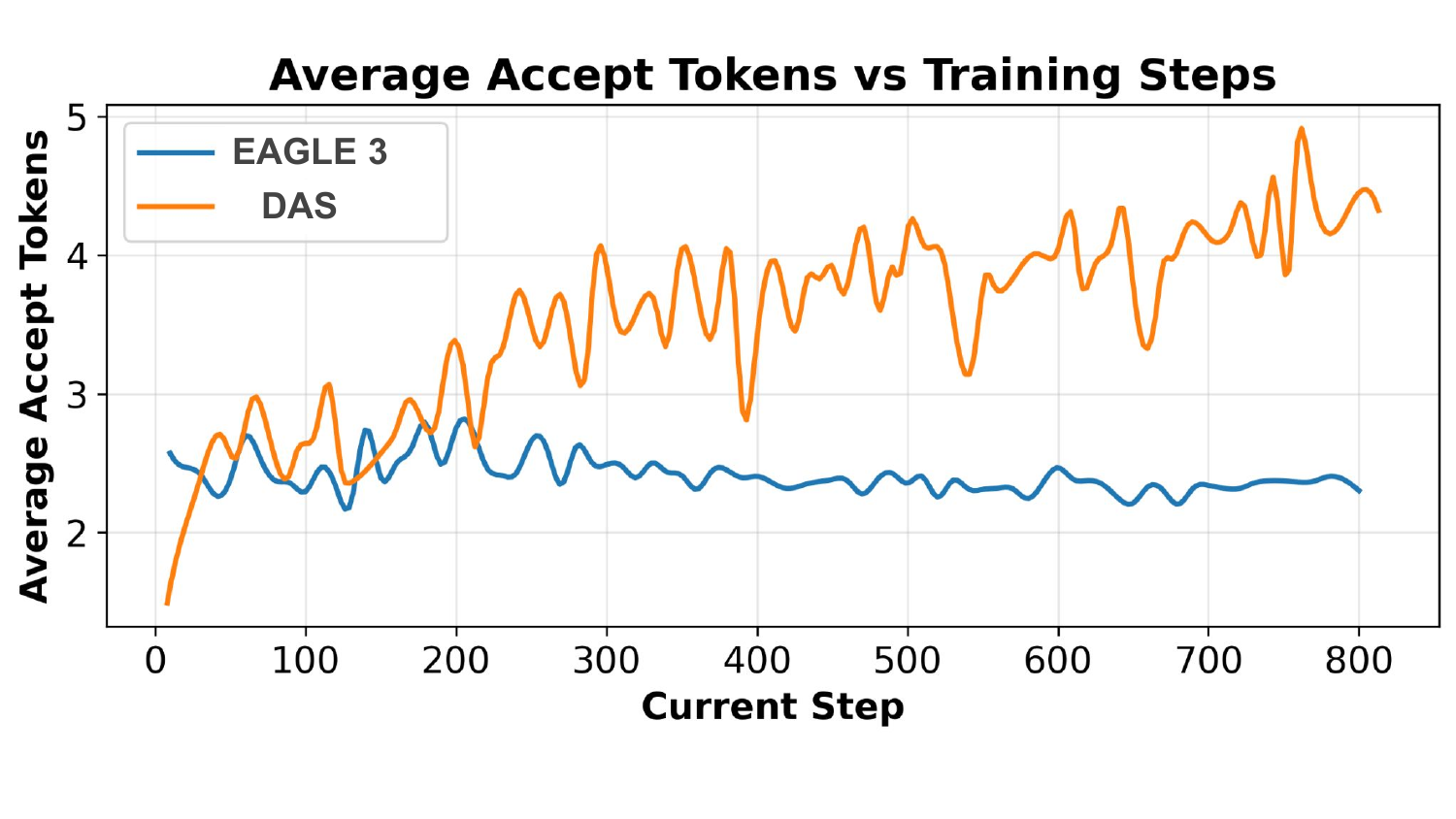}
    \caption{Average accepted tokens per verification round in RL training. We compare a static learned drafter , \textsc{EAGLE}~\citep{li2024eagle2} with our training free drafter. While the static drafter’s acceptance stays flat, our non-parametric drafter updates with recent rollouts and tracks the evolving policy, yielding higher accepted length over time.
Higher accepted length implies fewer target forward passes per generated token and thus lower rollout latency.}
    \label{fig:eagle}
\end{figure}

EAGLE~\citep{li2024eagle}, EAGLE-2~\citep{li2024eagle2} and EAGLE-3~\citep{li2025eagle3} are state-of-the-art inference-time serving techniques by speculative decoding. A lightweight EAGLE head extrapolates internal features to construct a token draft tree that the target model then verifies in parallel, yielding substantial latency reductions during inference~\citep{li2024eagle,li2024eagle2,li2025eagle3}. EAGLE-2 in particular depends on a well-calibrated drafter whose confidence closely predicts token acceptance, and it grows a dynamic draft tree accordingly~\citep{li2024eagle2}.

In RL training, however, the policy is non-stationary: model weights change after every learner update, so this calibration rapidly drifts. Figure~\ref{fig:generation_sim} implies that the trajectories collected from earlier-epoch models would lead to significantly lower acceptance rates than the trajectories collected from later checkpoints. 
In practice, EAGLE must either tolerate decreasing acceptance (and thus reduced speedup), or repeatedly re-train / re-tune the head and its tree-building thresholds throughout training — adding compute and engineering overhead to an already rollout-dominated stage.

In addition, EAGLE’s tree drafting and parallel verification kernels are engineered for high-throughput inference serving. Integrating and continuously re-calibrating these components inside an actor–learner RL pipeline, where checkpoints evolve over time, is considerably more complex than the intended inference-only use case. For these reasons, despite its strong inference results, EAGLE is difficult to scale efficiently in non-stationary RL training regimes.




The high overhead of continuously updating neural models motivates a model-free approach based on text indexes, specifically suffix trees~\citep{ukkonen1995line} and suffix arrays~\citep{manber1993suffix}.




\begin{figure}[ht]
    \centering
    \includegraphics[width=1\linewidth]{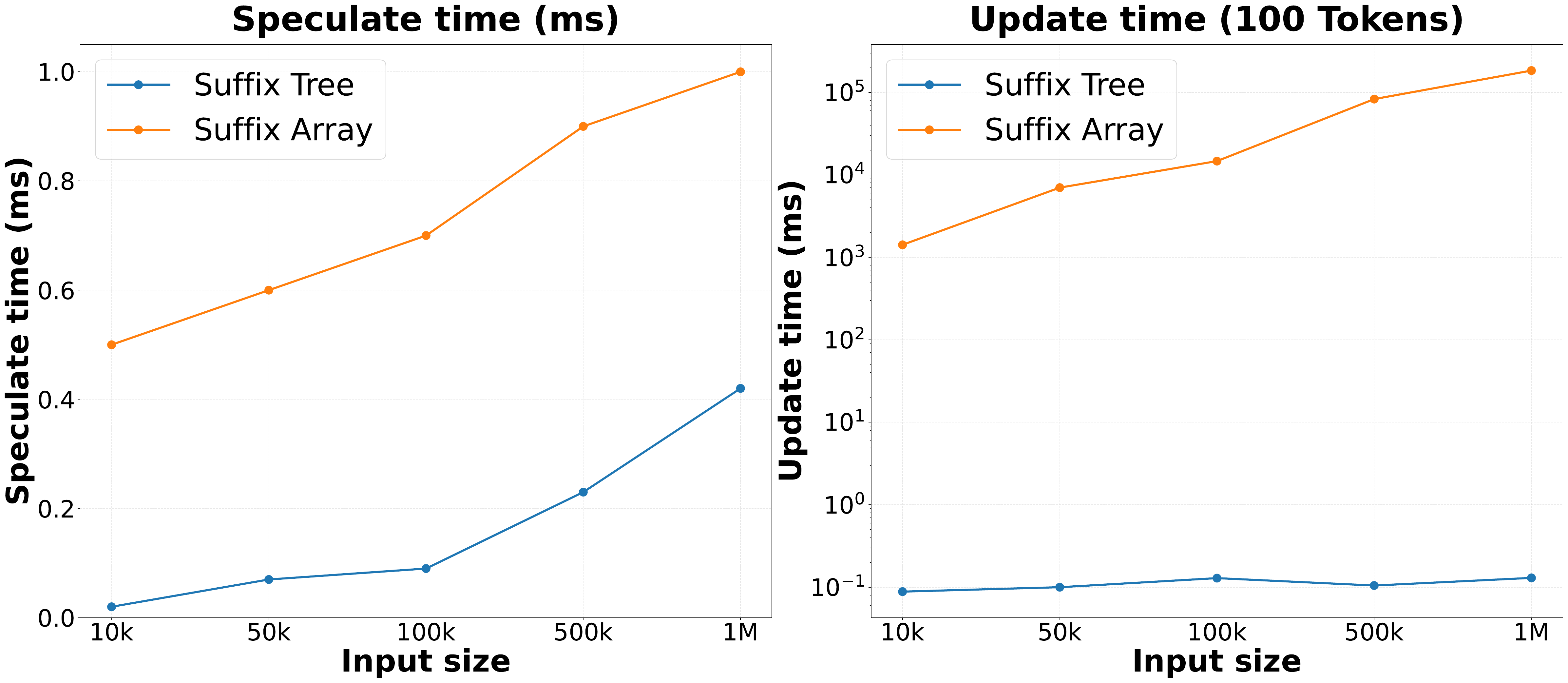}
    \caption{Performance comparison of suffix tree and suffix array data structures. Left: Speculation time across different corpus sizes. Right: Update time for inserting 100 tokens (log scale).}
    \label{fig:suffix}
\end{figure}


\subsubsection{Adaptive nonparametric drafter}
Building on the motivation above, we explore an \emph{adaptive nonparametric} drafter. We utilize a suffix-based drafter constructed from a sliding window of recent rollouts by maintaining a compact suffix tree over past generations. At rollout time, we locate the longest match between the current context and the tree (top-down in $O(m)$ time for a context of length $m$, where $m$ is the length of the query/pattern we look up---e.g., the current decode prefix), and then propose a multi-token draft along the matched path. The target model verifies this draft in one or a few batched steps. After verification, we update the tree online, so that the speculator continuously adapts to the evolving policy.

This design captures recurring motifs across training epochs while avoiding the need to train or maintain a separate training-based neural drafter. Importantly, the construction of the suffix-tree and updates of the online suffix-tree are run in linear time through the Ukkonen algorithm~\citep{ukkonen1995line}, which makes it suitable for the incremental intake of new routes.

Figure~\ref{fig:eagle} compares a parameterized static drafter (EAGLE) with our nonparametric adaptive drafter in RL training. We plot the \emph{average accepted tokens per verification round}. Although EAGLE maintains a roughly flat acceptance curve, the acceptance of our drafter continues to improve as training progresses, because it is continuously updated from recent rollouts.


\textbf{Suffix tree and suffix array.} In addition to the suffix tree, it is natural to utilize a suffix array (SA), inspired by Infini-gram \citet{infinigram}, as a more space-efficient alternative. A standard SA supports the search for substrings by binary search in $O(m \log n)$ time for a pattern of length $m$~\citep{manber1993suffix}, where $n$ is the number of tokens in the corpus. Augmenting the SA with an LCP array~\citep{kasai2001linear} reduces the number of comparisons, and using an enhanced suffix array (ESA)~\citep{abouelhoda2004replacing} enables pattern matching in $O(m)$ time by effectively simulating suffix-tree traversal with better constants and cache locality. However, SAs and ESAs are fundamentally static: dynamic updates typically require (partial) $O(n)$ rebuilding, which is undesirable in RL training, where fresh trajectories arrive every iteration.

Our online suffix-tree index instead trades modest additional space for fast incremental updates and $O(m)$ longest-match queries, which aligns naturally with the non-stationary policy setting.

We compare the speculative time and the update cost for the suffix tree versus the suffix array in Figure~\ref{fig:suffix}. The suffix tree demonstrates superior performance on both metrics: speculative times are 2-20× faster, while update costs show a dramatic advantage, remaining sub-millisecond compared to the suffix array's escalating reconstruction times. This over three orders of magnitude difference in update performance confirms that suffix arrays, despite their space efficiency, are impractical for online RL training, where fresh trajectories must be rapidly updated each iteration.






\begin{figure}[ht]
    \centering    \includegraphics[width=1\linewidth]{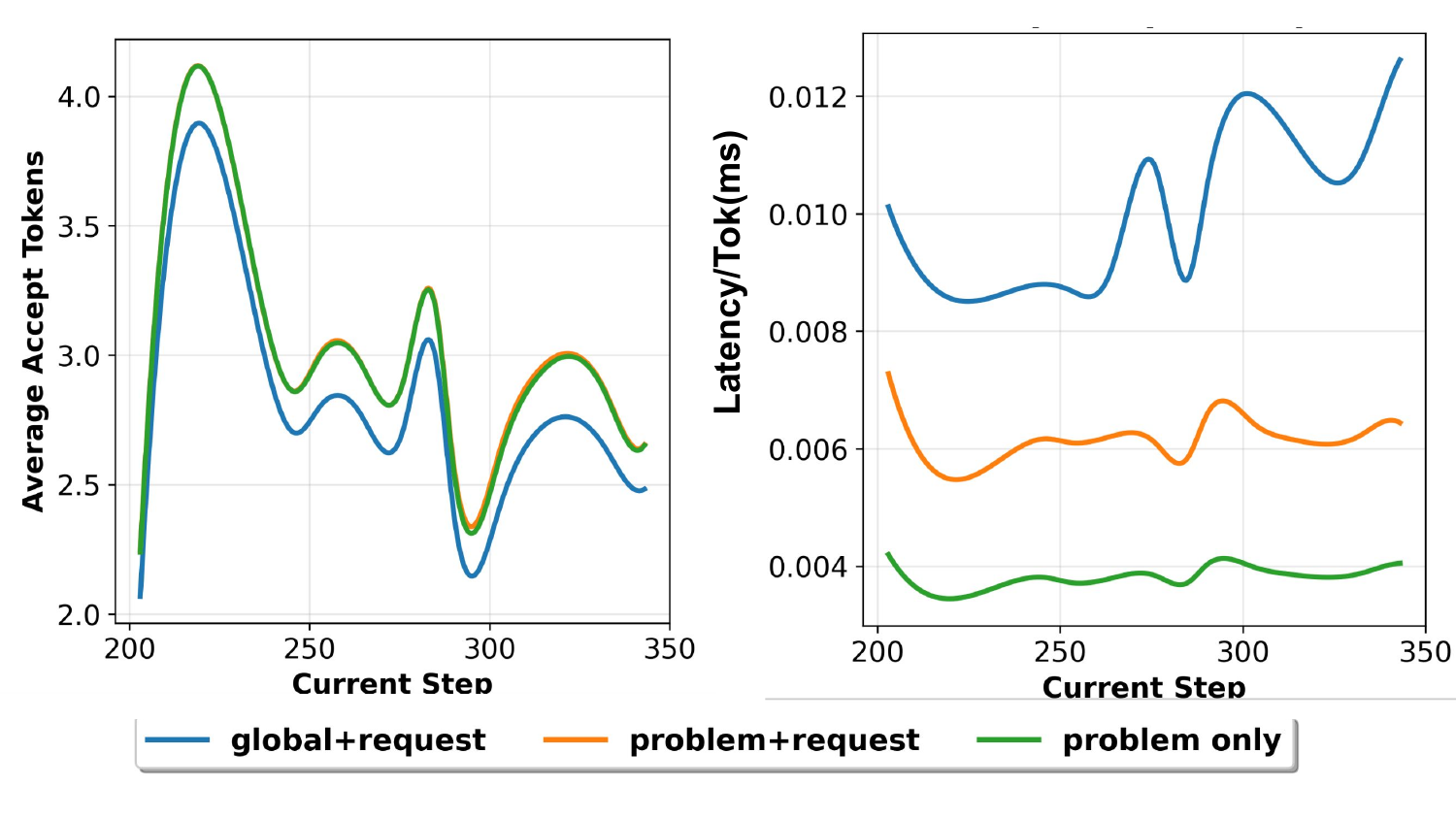}
    \caption{(\textbf{Left})~Average accepted tokens per verification round vs.\ training step. We compare three histories for building the drafter: \emph{global+request}, \emph{problem+request}, and \emph{problem only}; problem-scoped histories exceed global in acceptance. (\textbf{Right})~Speculative decoding latency (ms/token); \emph{global+request} is consistently slower due to the cost of querying and maintaining a single large global index, whereas problem-scoped shards stay cheaper to query and thus lower latency.}
    \label{fig:global_problem_prompt}
\end{figure}



\paragraph{Global suffix trees.} A single, ever-growing global index over all past rollouts fails along both statistical and systems axes. 
Statistically, policy drift causes older continuations to become poorly aligned with the model's current conditional distribution, which lowers speculative decoding acceptance rates — the very quantity that determines speedup (see Figure~\ref{fig:generation_sim}). Moreover, although all questions in the RL may lie within the same domain, their diversity means that patterns from one problem rarely transfer reliably to another. Consistent with this, Figure~\ref{fig:global_problem_prompt} shows that enabling a global tree yields smaller gains than maintaining a tree per-problem but increase additional overheads because of larger tree.







\paragraph{Per-request suffix trees.} 
Building on the above analysis, we adopt a \emph{per-request} suffix tree together with a lightweight pre-request prefix trie for routing. The key advantage of the per-problem design is that when the model tends to repeat a pattern, the prefix trie can quickly recognize that pattern and route to the corresponding suffix tree since it is constructed from prior generations. However, the benefit of prefix routing is workload and model-dependent. For smaller models, the additional CPU overhead of prefix routing can outweigh its gains. In such regimes, we disable the pre-request trie and query the per-problem suffix tree directly. Figure~\ref{fig:global_problem_prompt} shows that using a pre-request tree may lead to a higher acceptance rate but incurs more speculative time.

\paragraph{Sliding window selection tree.}
Across epochs and models, we observe substantial similarity among generations for the same prompts (Fig.~\ref{fig:generation_sim}). Nearby trajectories tend to produce longer accepted drafts, while trajectories from much earlier epochs yield shorter accepted lengths, indicating reduced match quality as the policy drifts.

Because policy evolves over time, long-history generations become less predictive. We therefore construct the drafter from a sliding window of recent trajectories and refresh the index for each iteration. The window size controls a bias–stability trade-off: shorter windows adapt quickly but offer fewer matches, whereas longer windows improve coverage but risk staleness. In practice, we tie the window update rate to the optimizer’s step scale (e.g., larger parameter updates imply shorter windows) and apply a mild down-weighting to matches originating from older epochs.

\begin{figure}[H]
    \centering    \includegraphics[width=1\linewidth]{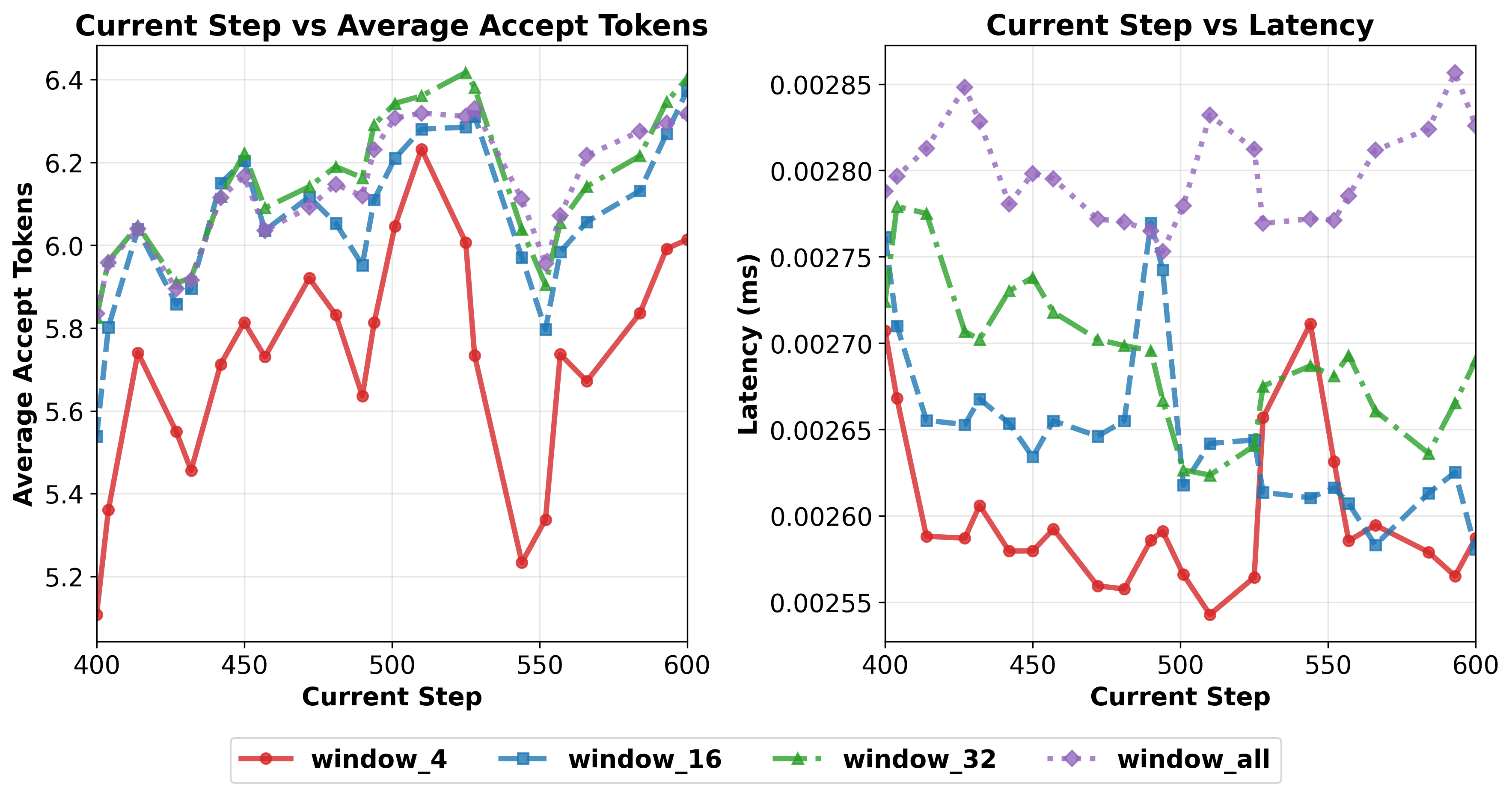}
    \caption{(\textbf{Left}) Average accepted tokens per verification round versus training step for different history window sizes used to build the drafter. Larger windows (e.g., 16, 32, all) give higher acceptance because they offer more matching continuations, and higher acceptance is known to translate directly into fewer target forward passes and lower decoding cost. (\textbf{Right}) Per-token speculative decoding latency versus training step. \texttt{window\_all} shows the highest latency because querying and maintaining a full global history is more expensive and includes stale trajectories, so in practice moderate windows (16 or 32) strike a better balance between acceptance and latency than \texttt{window\_all}.}
    \label{fig:generation_length}
\end{figure}

\subsection{length-aware Speculation Policy}
\label{sec:dynamic_draft_budget}


\subsubsection{Rollout Latency Estimation}
We estimate end-to-end rollout latency by first modeling the per–forward-pass latency and then accumulating it across all forward passes. From profiling, we find that a simple linear model is capable of capturing the main behavior (mean relative error $\approx 12\%$):
\begin{figure}[H]
    \centering
    \includegraphics[width=0.8\linewidth]{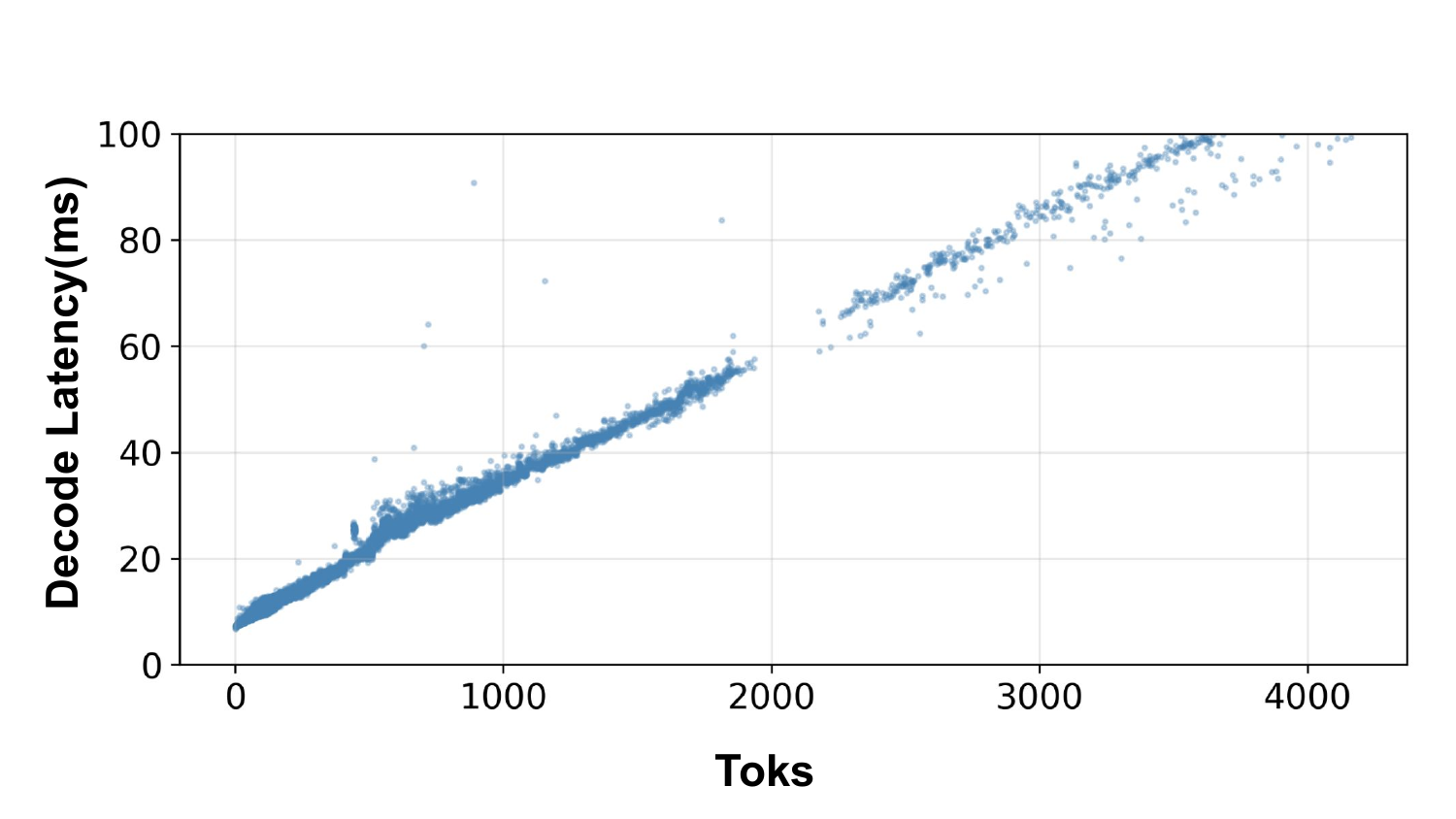}
    \caption{Decode latency vs.\ token number on DeepSeek-R1-Distill-Qwen-7B.  It shows clear linear relationship.}
    \label{fig:latency_estimation}
\end{figure}
\begin{equation}
t_{\mathrm{fwd}} \;=\; c_{\mathrm{base}} \;+\; c_{\mathrm{tok}} \, n_{\mathrm{toks}},
\end{equation}
where $n_{\mathrm{toks}}$ is the number of tokens processed (accepted + speculative).  
The term $c_{\mathrm{base}}$ captures per-pass overheads such as moving parameters/activations (global $\rightarrow$ shared memory), kernel launches, and temporary allocations.  
The term $c_{\mathrm{tok}}$ represents the average compute cost per-token on the GPU/CPU.

The total rollout latency is
\begin{equation}
t_{\mathrm{total}}
\;=\;
\sum_{i=1}^{N_{\mathrm{fwd}}} t_{\mathrm{fwd}} \;+\; C
\;=\;
c_{\mathrm{base}}\,N_{\mathrm{fwd}} \;+\; c_{\mathrm{tok}}\,N_{\mathrm{toks}} \;+\; C,
\end{equation}
where $C$ includes non-forward overheads such as input preparation, scheduling, and output formatting.

Thus, rollout latency decomposes into three parts:
(1) non-forward overhead $C$,
(2) base (per-pass) cost $c_{\mathrm{base}} N_{\mathrm{fwd}}$, and
(3) token-dependent cost $c_{\mathrm{tok}} N_{\mathrm{toks}}$.
To reduce latency, we should minimize both the number of forward passes $N_{\mathrm{fwd}}$ and the total number of decoded tokens $N_{\mathrm{toks}}$ (speculative and non-speculative).  
In speculative decoding, there is a trade-off: increasing speculative tokens can reduce $N_{\mathrm{fwd}}$, but proposing too many tokens can introduce extra system overhead (see Fig.~\ref{fig:latency_estimation}).  
An optimal strategy must therefore balance the speculative-token budget
to maximize overall rollout speedup.

This formulation also highlights why long generations dominate total latency:
they not only incur higher token-dependent cost, but also determine the number of forward passes required for the entire batch, thereby amplifying the base-cost component.

\subsubsection{Optimal Speculative-Token Budget}
To derive a good configuration for the speculative-token budget,
consider a batch of $n$ requests $\{r_i\}_{i=1}^n$, each characterized by:
\begin{itemize}
  \item Target generation length $l_i$,
  \item Accept efficiency parameter $\alpha_i > 0$,
  \item Total proposed token count $p_i$ (including speculative and non speculative tokens).
  \item drafter capacity factor
$k_i \in (0,1]$, which denotes the maximal achievable fraction of accepted tokens for request $r_i$.
\end{itemize}

The total number of accepted tokens follows a saturating form:
\begin{equation}
A_i(p_i) = k_i\, l_i \bigl(1 - e^{-\alpha_i p_i / l_i}\bigr),
\label{eq:acc_len_cap}
\end{equation}
where $A_i(p_i)\!\to\!k_i l_i$ as $p_i\!\to\!\infty$, 
reflecting the intrinsic mismatch limit between the two models.  
The remaining tokens to be generated are
\[
l_i - A_i(p_i) = l_i \bigl(1 - k_i + k_i e^{-\alpha_i p_i / l_i}\bigr),
\]
and the total number of forward passes needed to finish all requests is
\begin{equation}
N_{\mathrm{fwd}} = 
\max_i\, \bigl[l_i (1 - k_i + k_i e^{-\alpha_i p_i / l_i})\bigr].
\end{equation}

The corresponding rollout latency is modeled by
\begin{equation}
\begin{aligned}
J(\mathbf{p})
  &= c_{\mathrm{base}} \cdot 
    \max_i \bigl[l_i (1 - k_i + k_i e^{-\alpha_i p_i / l_i})\bigr] \\
  & \quad + c_{\mathrm{tok}} \cdot
    \sum_{i=1}^n p_i + C,
\label{eq:objective_cap}
\end{aligned}
\end{equation}
subject to $p_i \ge 0$ and any system-level constraints on speculative expansion.

Using $N_{\mathrm{fwd}}$ to denote the effective number of forward passes after speculation,
we can reformulate the optimization as
\begin{equation}
\begin{aligned}
\min_{\mathbf{p},\, N_{\mathrm{fwd}}} \quad
& c_{\mathrm{base}} N_{\mathrm{fwd}} + c_{\mathrm{tok}} \sum_i p_i,\\
\text{s.t.} \quad
& l_i (1 - k_i + k_i e^{-\alpha_i p_i / l_i}) \le N_{\mathrm{fwd}}, \;\; p_i \ge 0.
\end{aligned}
\label{eq:constrained_form_cap}
\end{equation}

At optimality, the constraint is tight for active requests:
\[
l_i (1 - k_i + k_i e^{-\alpha_i p_i / l_i}) = N_{\mathrm{fwd}},
\]
which gives a closed-form solution for the speculative budget:
\begin{equation}
\begin{aligned}
    p_i^*
  &= -\frac{l_i}{\alpha_i}
    \ln\!\Bigl(1 - k_i(1 - N_{\mathrm{fwd}}/l_i)\Bigr),
   \text{for } N_{\mathrm{fwd}} < l_i; \\
 p_i^* &= 0 \text{ otherwise.}
\end{aligned}
\label{eq:pi_star_cap}
\end{equation}

Substituting Eq.~\eqref{eq:pi_star_cap} into Eq.~\eqref{eq:objective_cap}
yields a single-variable objective:
\begin{equation}
\begin{aligned}
J(N_{\mathrm{fwd}})
  &= c_{\mathrm{base}} N_{\mathrm{fwd}} + C\\
   & \quad + c_{\mathrm{tok}}
      \sum_{i:\, l_i > N_{\mathrm{fwd}}}
      \frac{l_i}{\alpha_i}
      \Bigl[-\ln\!\bigl(1 - k_i(1 - N_{\mathrm{fwd}}/l_i)\bigr)\Bigr]
      . 
\end{aligned}
\label{eq:Jt_cap}
\end{equation}

Differentiating with respect to $N_{\mathrm{fwd}}$ gives the optimality condition:
\begin{equation}
 c_{\mathrm{base}}
    - c_{\mathrm{tok}}
      \sum_{i:\, l_i > N_{\mathrm{fwd}}}
      \frac{k_i}
           {\alpha_i(1 - k_i + k_i N_{\mathrm{fwd}} / l_i)}
  = 0.
\label{eq:opt_condition_cap}
\end{equation}

\paragraph{Observations.}
From this formulation, we make three key observations:
\begin{enumerate}
  \item The optimal speculative budget $p_i^*$ grows with the request length $l_i$, and requests with similar lengths receive similar speculative token budgets.
  \item Short generations with $l_i \le N_{\mathrm{fwd}}$ should skip speculation.
  \item The capacity factor $k_i$ bounds the maximum speculative gain.
        When $k_i$ is small (weak drafter), $p_i^*$ and the achievable speedup both shrink,
        as additional speculative tokens yield diminishing returns.
  \item When $c_{\mathrm{base}} \gg c_{\mathrm{tok}}$ (the base-cost-dominant regime),
        the optimal strategy prioritizes reducing the number of forward passes $N_{\mathrm{fwd}}$,
        consistent with empirical findings in small-batch rollout.
\end{enumerate}

These findings align with intuition: the more overhead a request incurs, the more compute effort should be allocated to it.  
Long generations contribute disproportionately to total latency and therefore warrant more aggressive speculative decoding,  
whereas short generations impose minimal overhead and thus benefit little from speculation.

\subsubsection{Dynamic Draft Budget via Runtime Length Prediction}
\begin{figure}[H]
    \centering    
    \includegraphics[width=1\linewidth]{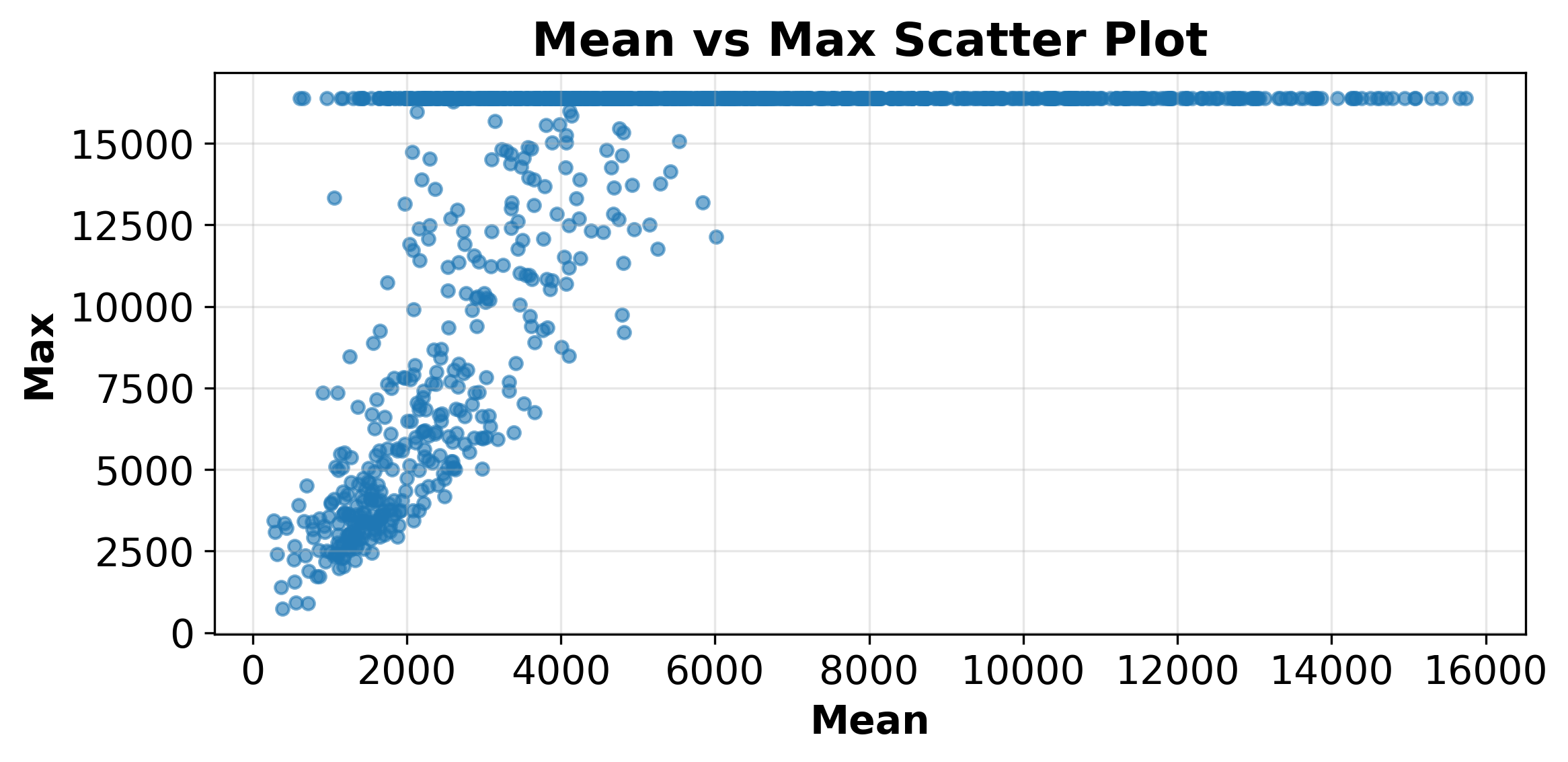}
    \caption{Each point represents one problem, where the x-axis shows the mean generation length across 90 epochs and the y-axis shows the maximum generation length observed. The wide spread and high upper bound indicate that generation length behavior is highly dynamic.}
    \label{fig:rollout length}
\end{figure}

From the previous section in step Eq~\ref{eq:opt_condition_cap}, we learned that accurate length prediction is crucial to selecting an appropriate budget draft. However, generation dynamics are highly stochastic (see Fig.~\ref{fig:rollout length}), making direct prediction difficult. We therefore propose a hierarchical heuristic that combines historical statistics with runtime signals:

\begin{enumerate}
  \item \textbf{Length-class policy.}
        We partition requests into three length classes—\emph{Long}, \emph{Medium}, and \emph{Short}—each mapped to a corresponding speculative budget.
        Long requests use a larger (more aggressive) speculative budget, Medium requests use an intermediate budget, and Short requests disable speculative decoding.

  \item \textbf{Initialization from history.}
        The initial class is chosen from the historical distribution for requests similar to $r$:
        \[
        \text{Init}_r
        \;=\;
        \arg\max_{c \in \{\text{Long},\text{Medium},\text{Short}\}}
        \#\{\,r' \sim r : r' \in c\,\},
        \]
        where $\#\{\cdot\}$ counts historical requests comparable to $r$ falling in class $c$.

  \item \textbf{Runtime update.}
        During generation, we update the class dynamically based on the observed partial length $l$:
        \[
        \text{Class}_r \mid l,\, \text{Init}_r
        \;=\;
        \arg\max_{c \in \{\text{Long},\text{Medium},\text{Short}\}}
        P\!\left(c \mid l,\, \text{Init}_r\right).
        \]
        We estimate $P\!\left(c \mid l,\, \text{Init}_r\right)$ from historical rollout statistics to obtain a practical prior for online length classification. The speculative decoding settings are then adjusted to match the current class.
\end{enumerate}






\section{End to End Speedup Experiments}









We evaluate our system on the SOTA RL training framework VeRL~\cite{sheng2024hybridflow}, which implements the core generation, reward, and training loop. 

We validate \Ours{} on two representative post-training RL workloads: mathematical reasoning and code generation.
For math, we follow recent One-Shot-RLVR~\citep{wang2025reinforcement} training recipes, where a policy is optimized for verifiable reasoning quality on competition-level problems using reinforcement learning and structured reward shaping. \Ours{} plugs into this loop without changing the reward model or optimizer.

For code, we adopt an RL setup similar to DeepCoder~\citep{deepcoder} pipelines: prompts specify a programming task, and reward is assigned by unit-test pass/fail of the generated program, a standard verifiable outcome signal in program synthesis and code RL.

In both cases, generation is data-parallel across actors, and speculation is only applied at decode time; the policy update step itself (e.g., GRPO ~\citep{guo2025deepseek} optimization in frameworks such as VeRL~\citep{sheng2024hybridflow}) is left unchanged. 

We scaled the experiments to models with parameters $\sim$14B and ran them on clusters of up to six 8xH100 nodes. Unless otherwise noted, all reported timing numbers are measured as actual wall-clock rollout step time, including batching, scheduling, and verification overhead.

\subsection{MATH RL}

\begin{figure}[H]
  \centering
  \begin{minipage}{0.49\linewidth}
    \centering
    \includegraphics[width=\linewidth]{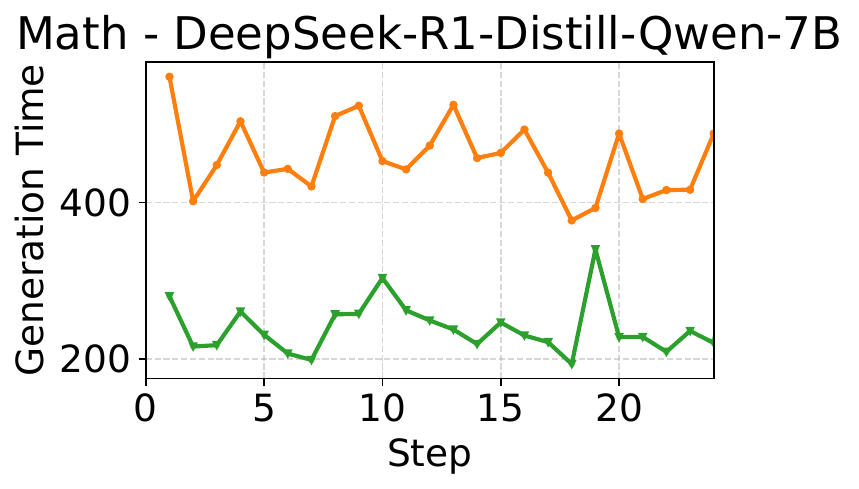}

  \end{minipage}\hfill
  \begin{minipage}{0.49\linewidth}
    \centering
    \includegraphics[width=\linewidth]{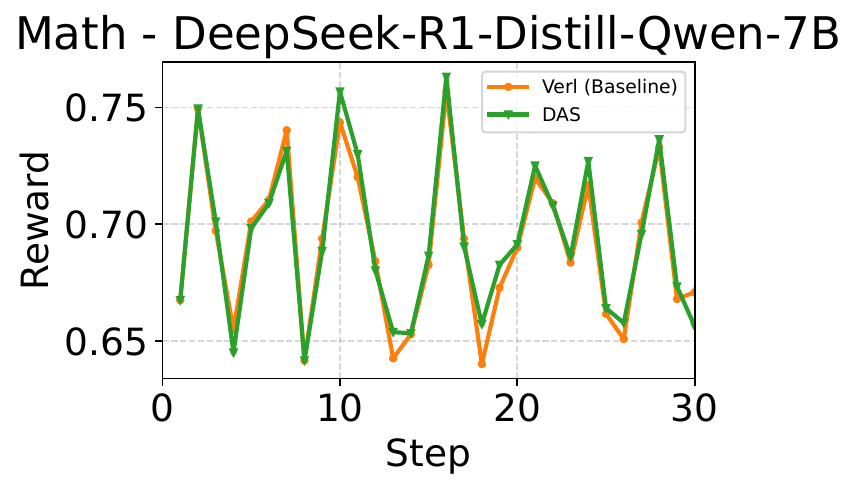}
  \end{minipage}
  \caption{
    Training curves for the \emph{DeepSeek-R1-Distill-Qwen-7B} model comparing the VeRL baseline (\textcolor{orange}{orange}) and our \Ours{}  approach (\textcolor{green}{green}).
    \textbf{Left:} Generation time (rollout latency) per training step. \Ours{} consistently reduces generation time by more than 50\% relative to the baseline.
    \textbf{Right:} Reward per training step. \Ours{}  closely matches the baseline reward across all steps, indicating no loss in training quality despite the speedup.
    }
  \label{fig:math_rl_reward}
\end{figure}

In Figure~\ref{fig:math_rl_reward}, we evaluate the DSR-sub dataset~\citep{wang2025reinforcement}, which consists of 1,209 examples from DeepScaleR~\citep{deepscaler2025}. We train with a maximum sequence length of 16K tokens, a training batch size of 128, and 16 samples per question on a single 8$\times$H100 node, yielding an effective batch size of $128 \times 16 / 8 = 256$. Training runs for 30 steps with a sampling temperature of $T = 0.6$. Because speculative decoding is a lossless acceleration approach aiming to preserve the rollout distribution, our \Ours{} system achieves an identical reward to the VeRL baseline. Our approach shows more than a 50\% reduction in total rollout time.

\subsection{Code RL}

We evaluate DeepCoder~\citep{deepcoder} on multi-step code execution with 8 samples per question, a per-GPU training batch size of 32, a maximum sequence length of 16K tokens, and a sampling temperature of $T = 0.6$. At each training step, the LLM generates code and executes it on a Ray cluster. The cluster consists of nodes with Intel(R) Xeon(R) Platinum 8468V CPUs. We configure Ray to schedule execution across all available CPU cores, ensuring that code execution is not bottlenecked by the environment.

\begin{figure}[H]
  \centering
  \begin{minipage}{0.49\linewidth}
    \centering
    \includegraphics[width=\linewidth]{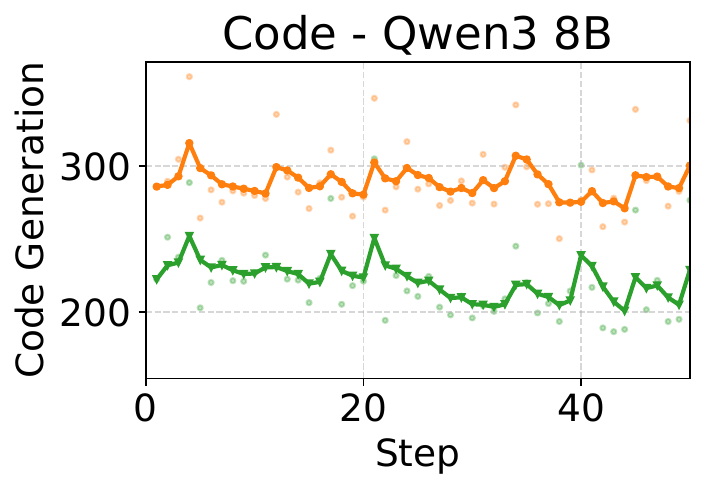}
  \end{minipage}\hfill
  \begin{minipage}{0.49\linewidth}
    \centering
    \includegraphics[width=\linewidth]{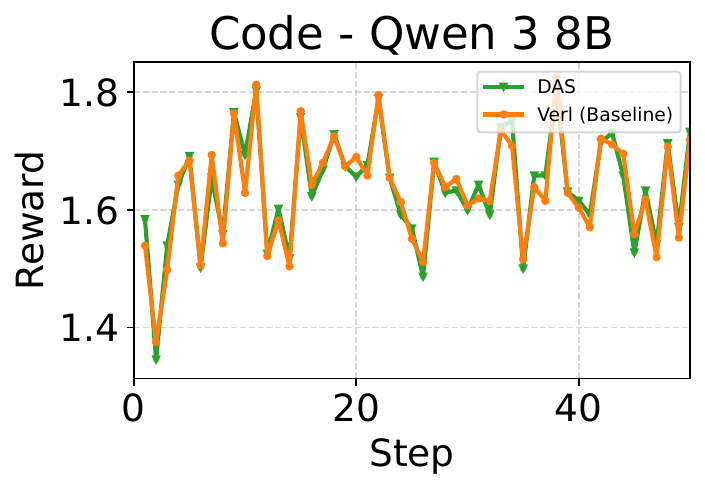}
  \end{minipage}
  \caption{
    Training curves for the \emph{Qwen3-8B} model comparing the VeRL baseline (\textcolor{orange}{orange}) and our \Ours{} approach (\textcolor{green}{green}).
    \textbf{Left:} Generation time (rollout latency) per training step. \Ours{} achieves substantially lower generation time throughout training.
    \textbf{Right:} Reward per training step. \Ours{} closely matches the baseline reward across steps, indicating no degradation in learning quality despite the speedup.
}
  \label{fig:code_rl_speed_reward}
\end{figure}

We train the \emph{Qwen3-8B} model on two 8$\times$H100 nodes using data parallelism. This yields an effective batch size of
$32 \times 8 / 16 = 16.$ In Figure~\ref{fig:code_rl_speed_reward}, we observe roughly a 25\% reduction in the rollout time while maintaining a comparable reward.

\subsection{Ablation Study}
\begin{figure}[H]
  \centering
  \begin{minipage}{0.49\linewidth}
    \centering
    \includegraphics[width=\linewidth]{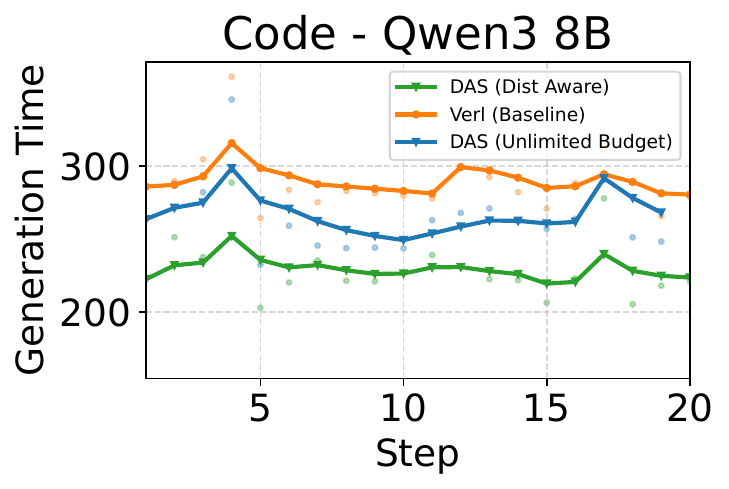}
  \end{minipage}\hfill
  \caption{Rollout generation time per training step for the \emph{Qwen3-8B} policy under three settings: the \textsc{VeRL} baseline (\textcolor{orange}{orange}), \Ours{} with an unlimited speculative budget (\textcolor{blue}{blue}), \Ours{} (\textcolor{green}{green}) with distribution aware. The unlimited budget variant (\textcolor{blue}{blue}) allows the drafter to propose arbitrarily long continuations, which increases verification cost and reduce the end-to-end gain by 15\% compared with distribution aware (\textcolor{green}{green}).
    }
  \label{fig:spec_budget}
\end{figure}

\textbf{Distribution Aware Speculative Decoding}
We evaluate the Distribution Aware Speculative decoding with the \emph{Qwen3-8B} Model. We provide a second baseline \emph{DAS Unlimited budget} that represents an unbounded speculate budget, enabling the suffix tree to propose as many tokens as possible. However, by proposing too many tokens, the cost of verification increases significantly, which reduces the potential improvement of speculative decoding. From Figure \ref{fig:spec_budget}, we can see that \Ours{} performs up to 15\% better than a budget-agnostic implementation.

\begin{figure}[H]
  \centering
  \begin{minipage}{0.49\linewidth}
    \centering
    \includegraphics[width=\linewidth]{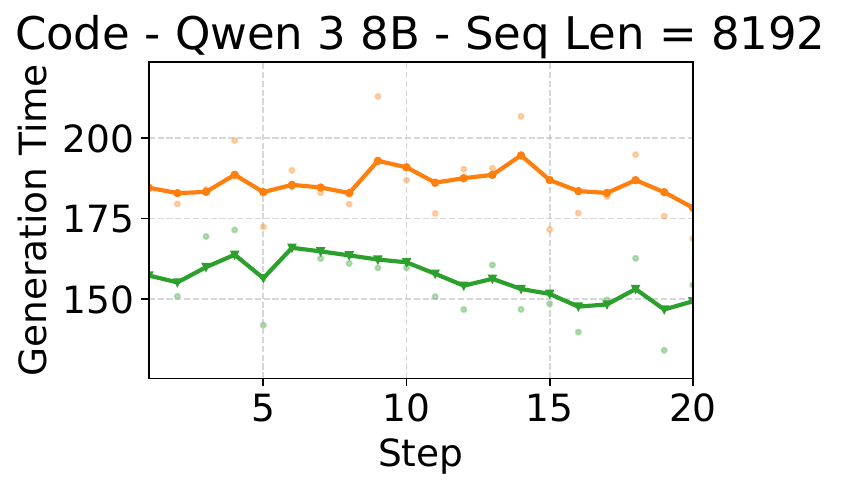}
  \end{minipage}\hfill
  \begin{minipage}{0.49\linewidth}
    \centering
    \includegraphics[width=\linewidth]{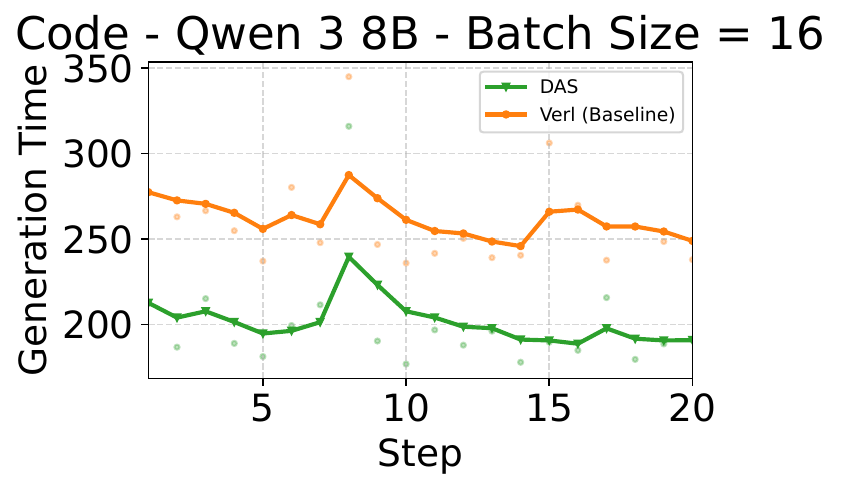}
  \end{minipage}
  \caption{
    Training curves for the \emph{Qwen3-8B} model comparing the VeRL baseline (\textcolor{orange}{orange}) and \Ours{} (\textcolor{green}{green}) under different sequence lengths (8k) and batch sizes (16).
    \textbf{Left:} Reducing the maximum generation length from 16k to 8k tokens still yields $>30\%$ end-to-end rollout speedup, indicating that \Ours{} continues to accelerate long, high-latency trajectories.
    \textbf{Right:} Reducing the effective batch size from 32 to 16 preserves a similar fractional speedup, showing that \Ours{} remains effective across different batch sizes. We omit the reward figure, as it was shown previously and follows from the distribution-preserving property of speculative decoding.
    }
  \label{fig:seq_bs_invariance}
\end{figure}

\textbf{Batch and Sequence Length.} We test robustness along two operational axes for \emph{Qwen3-8B} code training.
First, we adjust the maximum decode length from 8k to 16k tokens and still see over $30\%$ rollout speedup, confirming that our approach continues to attack the long-tail prompts that dominate makespan in RL rollout.
Second, we adjust the effective batch size from 32 to 16 and observe similar proportional savings, meaning our method does not rely on a specific batching regime to deliver latency reduction.
This invariance matters because speculative decoding delivers speed by cutting the number of sequential target-model forward passes per generated token, and that benefit should hold even as sequence length grows or batch size shrinks.


\section{Conclusion}

Reinforcement learning has become central to post-training large language models, but at current, scales most of the wall-clock cost comes from rollouts. Generating long trajectories is slow and the worst few samples in each batch set the step time.  Speculative decoding is a proven way to cut generation latency by having a fast drafter propose multiple tokens and letting the main model verify them in parallel without changing its output distribution. We adapt speculative decoding to RL by (i) replacing a fixed neural drafter with a non-parametric, adaptive drafter that is continuously refreshed from recent rollouts, and (ii) introducing a distribution-aware draft budgeting policy that allocates more speculative budget to harder, longer problems that dominate rollout time.  We validate this on both math reasoning and code RL settings, where rollouts are rewarded by verifiable signals (solution correctness or unit-test pass). In both cases, our method cuts rollout wall-clock time by over 30\% relative to a no-speculation baseline while matching the baseline reward curve, showing that adaptive speculative decoding can accelerate RL training without degrading policy quality.
\bibliography{reference}

@article{guo2025deepseek,
  title={Deepseek-r1: Incentivizing reasoning capability in llms via reinforcement learning},
  author={Guo, Daya and Yang, Dejian and Zhang, Haowei and Song, Junxiao and Zhang, Ruoyu and Xu, Runxin and Zhu, Qihao and Ma, Shirong and Wang, Peiyi and Bi, Xiao and others},
  journal={arXiv preprint arXiv:2501.12948},
  year={2025}
}

@article{sheng2024hybridflow,
  title   = {HybridFlow: A Flexible and Efficient RLHF Framework},
  author  = {Guangming Sheng and Chi Zhang and Zilingfeng Ye and Xibin Wu and Wang Zhang and Ru Zhang and Yanghua Peng and Haibin Lin and Chuan Wu},
  year    = {2024},
  journal = {arXiv preprint arXiv: 2409.19256}
}

@article{liu2024optimizing,
  title={Optimizing speculative decoding for serving large language models using goodput},
  author={Liu, Xiaoxuan and Daniel, Cade and Hu, Langxiang and Kwon, Woosuk and Li, Zhuohan and Mo, Xiangxi and Cheung, Alvin and Deng, Zhijie and Stoica, Ion and Zhang, Hao},
  journal={arXiv preprint arXiv:2406.14066},
  year={2024}
}

@article{huang2025specserve,
  title={Specserve: Efficient and slo-aware large language model serving with adaptive speculative decoding},
  author={Huang, Kaiyu and Wu, Hao and Shi, Zhubo and Zou, Han and Yu, Minchen and Shi, Qingjiang},
  journal={arXiv preprint arXiv:2503.05096},
  year={2025}
}

@article{infinigram,
  title={Infini-gram: Scaling unbounded n-gram language models to a trillion tokens},
  author={Liu, Jiacheng and Min, Sewon and Zettlemoyer, Luke and Choi, Yejin and Hajishirzi, Hannaneh},
  journal={arXiv preprint arXiv:2401.17377},
  year={2024}
}

@article{li2024eagle,
  title={Eagle: Speculative sampling requires rethinking feature uncertainty},
  author={Li, Yuhui and Wei, Fangyun and Zhang, Chao and Zhang, Hongyang},
  journal={arXiv preprint arXiv:2401.15077},
  year={2024}
}

@article{li2025eagle3,
  title={Eagle-3: Scaling up inference acceleration of large language models via training-time test},
  author={Li, Yuhui and Wei, Fangyun and Zhang, Chao and Zhang, Hongyang},
  journal={arXiv preprint arXiv:2503.01840},
  year={2025}
}

@article{li2024eagle2,
  title={Eagle-2: Faster inference of language models with dynamic draft trees},
  author={Li, Yuhui and Wei, Fangyun and Zhang, Chao and Zhang, Hongyang},
  journal={arXiv preprint arXiv:2406.16858},
  year={2024}
}

@article{oliaro2025suffixdecoding,
  title={SuffixDecoding: Extreme Speculative Decoding for Emerging AI Applications},
  author={Oliaro, Gabriele and Jia, Zhihao and Campos, Daniel and Qiao, Aurick},
  journal={arXiv preprint arXiv:2411.04975},
  year={2025}
}

@article{ukkonen1995line,
  title={On-line construction of suffix trees},
  author={Ukkonen, Esko},
  journal={Algorithmica},
  volume={14},
  number={3},
  pages={249--260},
  year={1995},
  publisher={Springer}
}

@article{manber1993suffix,
  title={Suffix arrays: a new method for on-line string searches},
  author={Manber, Udi and Myers, Gene},
  journal={siam Journal on Computing},
  volume={22},
  number={5},
  pages={935--948},
  year={1993},
  publisher={SIAM}
}

@article{abouelhoda2004replacing,
  title={Replacing suffix trees with enhanced suffix arrays},
  author={Abouelhoda, Mohamed Ibrahim and Kurtz, Stefan and Ohlebusch, Enno},
  journal={Journal of discrete algorithms},
  volume={2},
  number={1},
  pages={53--86},
  year={2004},
  publisher={Elsevier}
}

@article{hu2024openrlhf,
  title={Openrlhf: An easy-to-use, scalable and high-performance rlhf framework},
  author={Hu, Jian and Wu, Xibin and Shen, Wei and Liu, Jason Klein and Zhu, Zilin and Wang, Weixun and Jiang, Songlin and Wang, Haoran and Chen, Hao and Chen, Bin and others},
  journal={arXiv preprint arXiv:2405.11143},
  year={2024}
}

@inproceedings{leviathan2023fast,
  title={Fast inference from transformers via speculative decoding},
  author={Leviathan, Yaniv and Kalman, Matan and Matias, Yossi},
  booktitle={International Conference on Machine Learning},
  pages={19274--19286},
  year={2023},
  organization={PMLR}
}

@inproceedings{kasai2001linear,
  title={Linear-time longest-common-prefix computation in suffix arrays and its applications},
  author={Kasai, Toru and Lee, Gunho and Arimura, Hiroki and Arikawa, Setsuo and Park, Kunsoo},
  booktitle={Annual Symposium on Combinatorial Pattern Matching},
  pages={181--192},
  year={2001},
  organization={Springer}
}

@inproceedings{weiner1973linear,
  title={Linear pattern matching algorithms},
  author={Weiner, Peter},
  booktitle={14th Annual Symposium on Switching and Automata Theory (swat 1973)},
  pages={1--11},
  year={1973},
  organization={IEEE}
}

@misc{deepscaler2025,
  title={DeepScaleR: Surpassing O1-Preview with a 1.5B Model by Scaling RL},
  author={Michael Luo and Sijun Tan and Justin Wong and Xiaoxiang Shi and William Y. Tang and Manan Roongta and Colin Cai and Jeffrey Luo and Tianjun Zhang and Li Erran Li and Raluca Ada Popa and Ion Stoica},
  year={2025},
  howpublished={\url{https://pretty-radio-b75.notion.site/DeepScaleR-Surpassing-O1-Preview-with-a-1-5B-Model-by-Scaling-RL-19681902c1468005bed8ca303013a4e2}},
  note={Notion Blog}
}

@article{deepcoder,
  title={Deepcoder: A fully open-source 14b coder at o3-mini level},
  author={Luo, Michael and Tan, Sijun and Huang, Roy and Patel, Ameen and Ariyak, Alpay and Wu, Qingyang and Shi, Xiaoxiang and Xin, Rachel and Cai, Colin and Weber, Maurice and others},
  journal={Notion Blog},
  year={2025}
}

@article{miao2023specinfer,
   title={SpecInfer: Accelerating Large Language Model Serving with Tree-based Speculative Inference and Verification},
   journal={arXiv preprint arXiv:2305.09781v4},
   author={Miao, Xupeng and Oliaro, Gabriele and Zhang, Zhihao and Cheng, Xinhao and Wang, Zeyu and Zhang, Zhengxin and Wong, Rae Ying Yee and Zhu, Alan and Yang, Lijie and Shi, Xiaoxiang and Shi, Chunan and Chen, Zhuoming and Arfeen, Daiyaan and Abhyankar, Reyna and Jia, Zhihao},
   year={2023}}

@article{liu2024onlinesd,
      title={Online Speculative Decoding}, 
      author={Xiaoxuan Liu and Lanxiang Hu and Peter Bailis and Alvin Cheung and Zhijie Deng and Ion Stoica and Hao Zhang},
      year={2024},
      journal={arXiv preprint: arXiv:2310.07177}}

@article{xia2024swift,
      title={SWIFT: On-the-Fly Self-Speculative Decoding for LLM Inference Acceleration}, 
      author={Heming Xia and Yongqi Li and Jun Zhang and Cunxiao Du and Wenjie Li},
      year={2024},
      journal={arXiv preprint: arXiv:2410.06916}}

@inproceedings{zhang2024selfsd,
    title = "Draft {\&} Verify: Lossless Large Language Model Acceleration via Self-Speculative Decoding",
    author = "Zhang, Jun  and
      Wang, Jue  and
      Li, Huan  and
      Shou, Lidan  and
      Chen, Ke  and
      Chen, Gang  and
      Mehrotra, Sharad",
    booktitle = "Proceedings of the 62nd Annual Meeting of the Association for Computational Linguistics (Volume 1: Long Papers)",
    year={2024}
}

@article{liu2025specrl,
      title={SPEC-RL: Accelerating On-Policy Reinforcement Learning via Speculative Rollouts}, 
      author={Bingshuai Liu and Ante Wang and Zijun Min and Liang Yao and Haibo Zhang and Yang Liu and Anxiang Zeng and Jinsong Su},
      year={2025},
      journal={arXiv preprint: arXiv:2509.23232}
}

@article{zhang2025fastgrpo,
      title={FastGRPO: Accelerating Policy Optimization via Concurrency-aware Speculative Decoding and Online Draft Learning}, 
      author={Yizhou Zhang and Ning Lv and Teng Wang and Jisheng Dang},
      year={2025},
      journal={arXiv preprint: arXiv:2509.21792}
}

@article{he2025historyrhyme,
      title={History Rhymes: Accelerating LLM Reinforcement Learning with RhymeRL}, 
      author={Jingkai He and Tianjian Li and Erhu Feng and Dong Du and Qian Liu and Tao Liu and Yubin Xia and Haibo Chen},
      year={2025},
      journal={arXiv preprint: arXiv:2508.18588}
}

@article{wang2025reinforcement,
  title={Reinforcement learning for reasoning in large language models with one training example},
  author={Wang, Yiping and Yang, Qing and Zeng, Zhiyuan and Ren, Liliang and Liu, Liyuan and Peng, Baolin and Cheng, Hao and He, Xuehai and Wang, Kuan and Gao, Jianfeng and others},
  journal={arXiv preprint arXiv:2504.20571},
  year={2025}
}

@misc{verl_one_step_off_2024,
  title        = {Recipe: One Step Off Policy Async Trainer},
  author       = {{Verl Contributors}},
  year         = {2024},
  howpublished = {\url{https://verl.readthedocs.io/en/latest/advance/one_step_off.html}},
  note         = {Accessed: 2025-10-30},
  publisher    = {Read the Docs},
  url          = {https://verl.readthedocs.io/en/latest/advance/one_step_off.html}
}

@misc{zhong2025streamrlscalableheterogeneouselastic,
      title={StreamRL: Scalable, Heterogeneous, and Elastic RL for LLMs with Disaggregated Stream Generation}, 
      author={Yinmin Zhong and Zili Zhang and Xiaoniu Song and Hanpeng Hu and Chao Jin and Bingyang Wu and Nuo Chen and Yukun Chen and Yu Zhou and Changyi Wan and Hongyu Zhou and Yimin Jiang and Yibo Zhu and Daxin Jiang},
      year={2025},
      eprint={2504.15930},
      archivePrefix={arXiv},
      primaryClass={cs.LG},
      url={https://arxiv.org/abs/2504.15930}, 
}

@inproceedings {zhong2024distserve,
author = {Yinmin Zhong and Shengyu Liu and Junda Chen and Jianbo Hu and Yibo Zhu and Xuanzhe Liu and Xin Jin and Hao Zhang},
title = {{DistServe}: Disaggregating Prefill and Decoding for Goodput-optimized Large Language Model Serving},
booktitle = {18th USENIX Symposium on Operating Systems Design and Implementation (OSDI 24)},
year = {2024},
isbn = {978-1-939133-40-3},
address = {Santa Clara, CA},
pages = {193--210},
url = {https://www.usenix.org/conference/osdi24/presentation/zhong-yinmin},
publisher = {USENIX Association},
month = jul
}

@inproceedings{ouyang2022training,
 author = {Ouyang, Long and Wu, Jeffrey and Jiang, Xu and Almeida, Diogo and Wainwright, Carroll and Mishkin, Pamela and Zhang, Chong and Agarwal, Sandhini and Slama, Katarina and Ray, Alex and Schulman, John and Hilton, Jacob and Kelton, Fraser and Miller, Luke and Simens, Maddie and Askell, Amanda and Welinder, Peter and Christiano, Paul F and Leike, Jan and Lowe, Ryan},
 booktitle = {Advances in Neural Information Processing Systems},
 editor = {S. Koyejo and S. Mohamed and A. Agarwal and D. Belgrave and K. Cho and A. Oh},
 pages = {27730--27744},
 publisher = {Curran Associates, Inc.},
 title = {Training language models to follow instructions with human feedback},
 volume = {35},
 year = {2022}
}

@misc{bai2022constitutionalaiharmlessnessai,
      title={Constitutional AI: Harmlessness from AI Feedback}, 
      author={Yuntao Bai and Saurav Kadavath and Sandipan Kundu and Amanda Askell and Jackson Kernion and Andy Jones and Anna Chen and Anna Goldie and Azalia Mirhoseini and Cameron McKinnon and Carol Chen and Catherine Olsson and Christopher Olah and Danny Hernandez and Dawn Drain and Deep Ganguli and Dustin Li and Eli Tran-Johnson and Ethan Perez and Jamie Kerr and Jared Mueller and Jeffrey Ladish and Joshua Landau and Kamal Ndousse and Kamile Lukosuite and Liane Lovitt and Michael Sellitto and Nelson Elhage and Nicholas Schiefer and Noemi Mercado and Nova DasSarma and Robert Lasenby and Robin Larson and Sam Ringer and Scott Johnston and Shauna Kravec and Sheer El Showk and Stanislav Fort and Tamera Lanham and Timothy Telleen-Lawton and Tom Conerly and Tom Henighan and Tristan Hume and Samuel R. Bowman and Zac Hatfield-Dodds and Ben Mann and Dario Amodei and Nicholas Joseph and Sam McCandlish and Tom Brown and Jared Kaplan},
      year={2022},
      eprint={2212.08073},
      archivePrefix={arXiv},
      primaryClass={cs.CL},
      url={https://arxiv.org/abs/2212.08073}, 
}

@inproceedings{rlhfuse_2025,
author = {Zhong, Yinmin and Zhang, Zili and Wu, Bingyang and Liu, Shengyu and Chen, Yukun and Wan, Changyi and Hu, Hanpeng and Xia, Lei and Ming, Ranchen and Zhu, Yibo and Jin, Xin},
title = {Optimizing RLHF training for large language models with stage fusion},
year = {2025},
isbn = {978-1-939133-46-5},
publisher = {USENIX Association},
address = {USA},
booktitle = {Proceedings of the 22nd USENIX Symposium on Networked Systems Design and Implementation},
articleno = {26},
numpages = {15},
location = {Philadelphia, PA, USA},
series = {NSDI '25}
}

@inproceedings{fast_inference_via_spec_decoding,
author = {Leviathan, Yaniv and Kalman, Matan and Matias, Yossi},
title = {Fast inference from transformers via speculative decoding},
year = {2023},
publisher = {JMLR.org},
booktitle = {Proceedings of the 40th International Conference on Machine Learning},
articleno = {795},
numpages = {13},
location = {Honolulu, Hawaii, USA},
series = {ICML'23}
}

@inproceedings{MLSYS2024_42a452cb,
 author = {Lin, Ji and Tang, Jiaming and Tang, Haotian and Yang, Shang and Chen, Wei-Ming and Wang, Wei-Chen and Xiao, Guangxuan and Dang, Xingyu and Gan, Chuang and Han, Song},
 booktitle = {Proceedings of Machine Learning and Systems},
 editor = {P. Gibbons and G. Pekhimenko and C. De Sa},
 pages = {87--100},
 title = {AWQ: Activation-aware Weight Quantization for On-Device LLM Compression and Acceleration},
 volume = {6},
 year = {2024}
}

@inproceedings{kwon2023efficient,
  title={Efficient Memory Management for Large Language Model Serving with PagedAttention},
  author={Woosuk Kwon and Zhuohan Li and Siyuan Zhuang and Ying Sheng and Lianmin Zheng and Cody Hao Yu and Joseph E. Gonzalez and Hao Zhang and Ion Stoica},
  booktitle={Proceedings of the ACM SIGOPS 29th Symposium on Operating Systems Principles},
  year={2023}
}
\bibliographystyle{mlsys2026}

\appendix
\section*{Appendix C. Derivation of the Nonlinear Acceptance Model}
\paragraph{Per-round Acceptance Dynamics.}
\label{acc_cal}
During speculative decoding, each request $r_i$ proceeds through several draft rounds.
Let $d_{i,k}$ denote the number of speculative tokens proposed in round $k$,
and $a_{i,k}$ the acceptance rate in that round.
Empirical traces (e.g., \textsc{DySpec}, \textsc{DeepSeek-R1})
show that acceptance decays exponentially with round depth:
\begin{equation}
a_{i,k} = a_{i,0} e^{-\beta_i (k-1)},
\end{equation}
where $\beta_i$ quantifies how quickly the draft–main-model mismatch grows.

\paragraph{Aggregating over Rounds.}
The total number of accepted tokens after $K_i$ rounds is
\begin{equation}
A_i = \sum_{k=1}^{K_i} a_{i,k} d_{i,k}
     = a_{i,0} d_i \frac{1 - e^{-\beta_i K_i}}{1 - e^{-\beta_i}},
\end{equation}
assuming a fixed proposal length $d_i$ per round ($p_i = K_i d_i$).
Substituting $K_i = p_i/d_i$ gives
\begin{equation}
A_i(p_i)
  = \frac{a_{i,0} d_i}{1 - e^{-\beta_i}}
    \bigl(1 - e^{-\beta_i p_i / d_i}\bigr).
\end{equation}

\paragraph{Normalization and Simplification.}
For simplicity we re-parameterize constants
to express the same exponential-saturation behavior in normalized form:
\begin{equation}
A_i(p_i) = k_il_i \bigl(1 - e^{-\alpha_i p_i / l_i}\bigr),
\end{equation}
where $\alpha_i$ encodes the effective "draft efficiency" of request $i$, and 
$k_i \in (0,1]$ denotes the maximal achievable fraction of accepted tokens for request $r_i$.
When $p_i \ll l_i$, the model reduces to a linear regime
$A_i \approx \alpha_i p_i$;
when $p_i \gg l_i/\alpha_i$, it saturates at $A_i \!\to\! l_i$.





\end{document}